\documentclass[11pt,a4paper]{article}
\usepackage[hyperref]{acl2021}
\usepackage{times}
\usepackage{latexsym}
\usepackage{paralist}

\usepackage{microtype}
\usepackage{amsmath}
\aclfinalcopy 

\usepackage[utf8]{inputenc}
\usepackage[T1]{fontenc}
\usepackage[english]{babel}
\usepackage{amsmath}
\usepackage{amsfonts}
\usepackage{amssymb}
\usepackage{graphicx}
\usepackage{float}
\usepackage{multicol}
\usepackage{booktabs}
\usepackage{multirow}
\title{ Masked Adversarial Generation for Neural Machine Translation}

\author{Badr Youbi Idrissi\thanks{worked conducted during an internship at Naver Labs Europe} \\
 Centrale Supelec \\
  \texttt{\small badryoubiidrissi@gmail.com} \\\And
  Stéphane Clinchant \\
  Naver Labs Europe  \\
  \texttt{\small stephane.clinchant@naverlabs.com} \\}

\begin{document}
\maketitle
    

	\begin{abstract}
    Attacking Neural Machine Translation models is an inherently combinatorial task on discrete sequences, solved with approximate heuristics. Most methods use the gradient to attack the model on each sample independently.
    Instead of mechanically applying the gradient, could we learn to produce meaningful adversarial attacks ?
    In contrast to existing approaches, we learn to attack a model by training an adversarial generator based on a language model. We propose the Masked Adversarial Generation (MAG) model, that learns to perturb the translation model throughout the training process. The experiments show that it improves the robustness of machine translation models, while being faster than competing methods.
	\end{abstract}
	
	\section{Introduction}
Do neural machine translation models generalize well, or are they brittle to simple changes that a human would consider natural?
The seminal work of  \cite{szegedy2014} showed that deep neural networks were prone to adversarial attacks: by changing a few pixels in an image, the classifier's decision could be changed arbitrary. Similarly, Neural Machine Translation (NMT) models have been shown to be very sensitive not only to both natural noise (typos, repetitions etc...) and synthetic noise (character swaps, ...)\cite{belinkov2018} but also to domain shift:  \cite{koehn2017,muller2019}. Finally, they suffer from exposure bias, one of the causes of hallucinations as advocated by \cite{wang2020}.

In this paper, we propose a new model, Masked Adversarial Generation (MAG), which learns an adversarial generator to improve robustness for machine translation models. This paper is structured as follows. In section \ref{sec:soa}, we review the literature. Section \ref{sec:mag} we discuss our model
and proceed to experiments.


\section{Adversarial Training}
\label{sec:soa}

One potential solution to robustness issues is data augmentation: training the model on noisy text generated from the training corpus while keeping the same reference. While this method is efficient for noise that is easily generated \cite{karpukhin2019}, it generalizes poorly to new types of noise.

A second solution would be to employ gradient based attacks due to their preeminent use in computer vision.
For instance \cite{sato2019} applies virtual adversarial training \cite{miyato2016} to NMT. On the other hand,
HotFlip and Adversarial Examples for Character Level NMT \cite{ebrahimi2018,ebrahimi2018a} were the first to our knowledge to propose a discrete white box adversarial method based on gradients. 
Seq2Sick \cite{cheng2020} is a carefully crafted projected gradient attack on sequence to sequence models.
However, 200 steps are used in their experiments, rendering the attack unusable for adversarial training.

\cite{michel2019} criticize the previous unconstrained adversarial attacks and argue that an adversarial attack should preserve the meaning of sentences. Therefore,  they limit the perturbations to the k nearest word embedding neighbors or with a charswap operation on few words.
Similarly to \cite{ebrahimi2018} and \cite{cheng2020}, they use a greedy approach to generate multiple word substitutions. The greedily calculated attacks significantly decrease translation quality.

\cite{cheng2018,cheng2019,cheng2020a} propose various adversarial training methods that drastically improve the performance of NMT models.  In their doubly adversarial inputs paper, meaning preserving perturbations  are generated thanks to a language model in order to restrict the set of plausible substitutions. Their latest paper AdvAug \cite{cheng2020a} combined the doubly adversarial inputs with mixup regularization. 
Finally, \cite{acl_rl_advgen} advocated the use of Reinforcement Learning to generate adversarial examples.

Most gradient based methods above train robust neural machine translation models with adversarial attacks that are one-shot (all perturbations computed in parallel and independently for a given sentence). They also mostly use a single gradient ascent step. 
Greedy and Iterative attacks are more powerful but create a bottleneck since they require sequential computations. Therefore, they are inconvenient for most MT tasks due to their high computational cost.
\subsection{Doubly Adversarial Inputs}
The Doubly Adversarial Inputs \cite{cheng2019} paper was the starting point of this work: 
Let $\mathbf{x}=(x_1,\dots,x_s)$ be the input source sentence tokens, and $\mathbf{y}=(y_1,\dots,y_t)$ the input target sentence tokens.
Let $\hat{\mathbf{x}},\hat{\mathbf{y}}$ be the perturbed sentences.
Doubly adversarial inputs chooses 15\% of the tokens in the source sentence randomly. For each token $x_i$ in this set, the replacement $\hat{x_i}$ is computed in a subset of the vocabulary $\mathcal{V}_{x_i} \subset \mathcal{V}$.
\begin{equation*}
    \hat{x_i} = \text{arg}\max_{x\in \mathcal{V}_{x_i} } \cos(\mathrm{e}(x)-e_i, , \nabla_{e_i} \mathcal{L}(\mathbf{x},\mathbf{y}))
\end{equation*}where $e$ is the token embedding function. And $\mathcal{L}(\mathbf{x},\mathbf{y})$ is the standard cross entropy loss computed at $\mathbf{x}$ and $\mathbf{y}$. \cite{cheng2019} constrained the attack with meaning preserving tokens. To achieve this, the subset $\mathcal{V}_{x_i}$ is the top $n$ predictions of a bidirectional language model. 
Let $Q\left(x_{i}, \mathbf{x}\right) \in \mathbb{R}^{|\mathcal{V}|}$ be the likelihood of the $i$-th word in the sentence $x$.
They define $\mathcal{V}_{x_{i}}=\operatorname{top_n}\left(Q\left(x_{i}, \mathbf{x}\right)\right)$ as the set of the $n$ most
probable words in terms of the language model $Q\left(x_{i}, \mathbf{x}\right)$.

The perturbations are computed independently and correspond therefore to a one-shot attack for the encoder.
Once the total source sentence perturbation $\hat{\mathbf{x}}$ is calculated, the loss is computed using the perturbed source sentence $\mathcal{L}(\hat{\mathbf{x}}, \mathbf{y})$ and $\hat{y}$  is generated similarily to $\mathbf{\hat{x}}$.
Thus, the algorithm is doubly adversarial, since it attacks the model on the source then on the target side. The doubly adversarial loss corresponds to $\mathcal{L}(\hat{\mathbf{x}}, \hat{\mathbf{y}})$. The training is then performed by minimizing the following loss:
\begin{equation}
\min_{\theta} \mathcal{L}(\mathbf{x}, \mathbf{y};\theta )+ \mathcal{L}( \hat{\mathbf{x}}, \hat{\mathbf{y}};\theta )
\end{equation}

\section{Masked Adversarial Generation}\label{sec:mag}

Let us now turn our attention to a more general question: \textit{Can we directly learn an adversarial generator ?} Instead of picking substitute words \textit{a posteriori} thanks to the gradient information, could we design a generator that would be trained to maximize translation loss?

The key idea here is to consider a Masked Language Model (MLM) as a generator (cf \cite{clark2019}). The input is passed through the MLM that modifies it and then feeds it to the translation model. This generator learns the MLM task, while also generating data that would break the translation model on top. In other words, we would like the generator to produce substitutions which are very likely according to a language model but would yield a poor translation. To do so, we use the Gumbel-Softmax reparametrization\cite{jang2016,MaddisonMT17_concrete}. The Masked
Adversarial Generation Model (MAG) is defined by the following equations:
\begin{figure}[hbt!]
    \centering
    \includegraphics[width=0.45\textwidth]{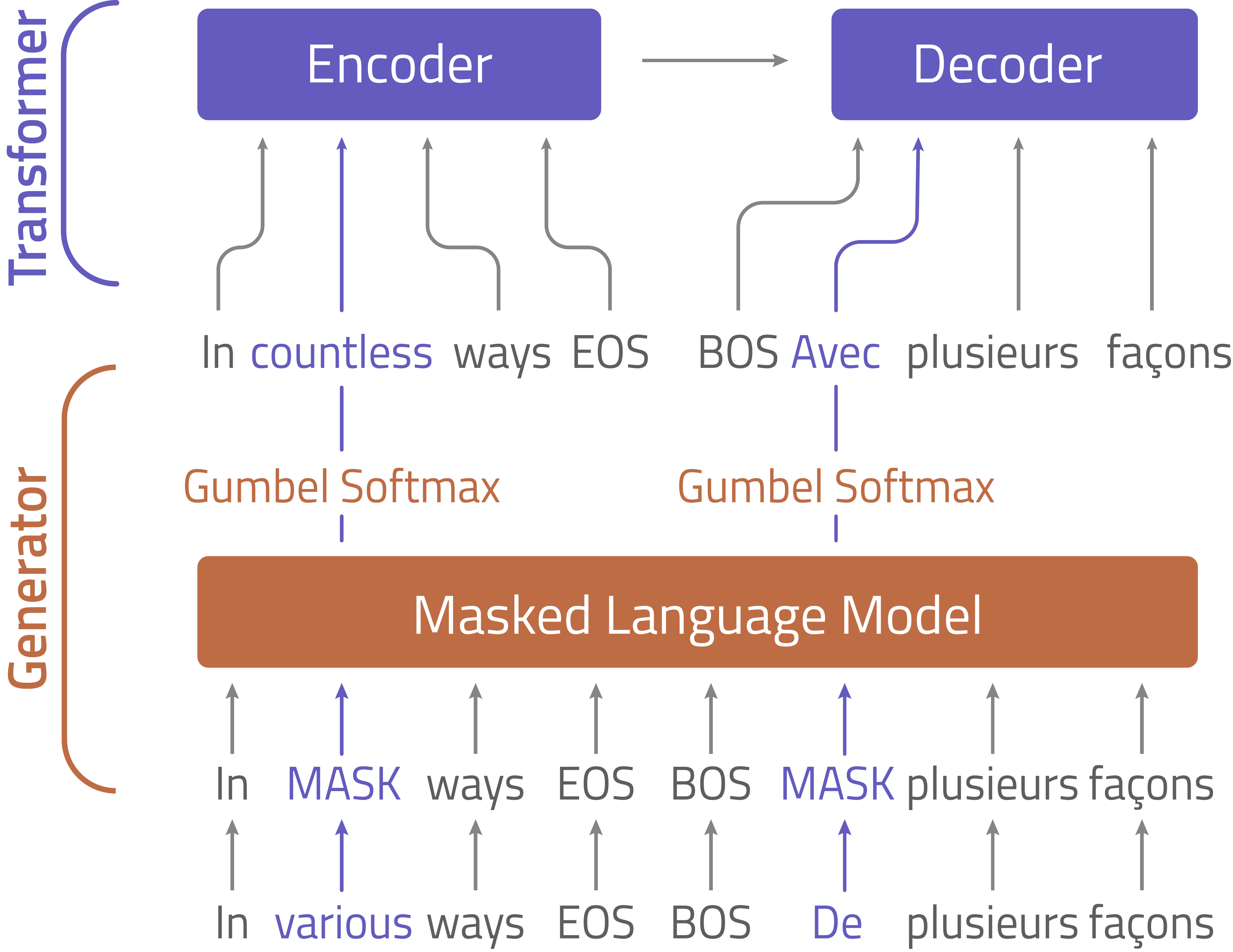}
    \caption{Masked Adversarial Generation Model}
    \label{fig:mag}
\end{figure}

\begin{eqnarray}
\hat{\mathbf{x}},\hat{\mathbf{y}} &\sim &  \text{GumbelSoftmax}( \text{mlm}[xy] ) \nonumber \\
 \min_{\theta} & & \mathcal{L}_{mt}(\mathbf{x}, \mathbf{y};\theta)+ \mathcal{L}_{mt}( \hat{\mathbf{x}}, \hat{\mathbf{y}};\theta ) + \mathcal{L}_{discr}(\hat{\mathbf{x}}, \hat{\mathbf{y}};\theta) \nonumber \\
  \min_{\phi} & & \mathcal{L}_{mlm}([\mathbf{x}\mathbf{y}];\phi) -\gamma  \mathcal{L}_{mt}( \hat{\mathbf{x}}, \hat{\mathbf{y}}; \phi )
             \nonumber
\end{eqnarray}
$[\mathbf{x}\mathbf{y}]$ refer to the concatenation of source and target sentences. $\hat{\mathbf{x}},\hat{\mathbf{y}}$ are sampled from the output of the joint MLM, similar to XLM \cite{xlm}, but using the gumbel softmax. Compared to a pair of bidirectional models, the joint MLM is faster and simpler, and acts as a simple non autoregressive translation model. $\gamma$ influences the  tradeoff between the adversarial loss and MLM loss for the generator. $\mathcal{L}_{discr}$ is an optional \textit{discrimination} loss  to detect replaced tokens as done in ELECTRA \cite{clark2019}.

Both the generator and translation model are learned from scratch and concurrently trained. We leave the study of pretrained generators to future work. Compared to the doubly adversarial inputs training algorithm, our proposal is much less computationally taxing. It only performs a single forward and backward pass during training.
In contrast, the doubly adversarial inputs approach uses 2 bidirectional language models that are roughly the same size as the translation model and does 2 backward and forward passes.
In our reimplementation, the doubly adversarial algorithm is 3.5 times slower than a normal transformer vs 1.5 for our model on the WMT17 dataset.

	\section{Experiments}
\subsection{Datasets and Setup}

We evaluated our reimplementation of the doubly adversarial model and MAG on two datasets. WMT17 English to German (with 4.5M sentences) and IWSLT 14 German to English (de-en with 0.2M). 
We use \textit{fairseq} \cite{ott2019fairseq} for the experiments and all metrics are computed with \textit{sacrebleu} \cite{sacrebleu}. All the models use the default Transformer-base \cite{vaswani2017} architecture hyperparameters for the translation task.
The learning rate schedule, decoding hyperparameters, and batch size are also the same. WMT17 experiments are run on 4 V100 gpus with 200K steps. In doubly adversarial inputs, the bidirectional language models are in total twice the size of the original Transformer. As in ELECTRA \cite{clark2019}, the MAG generator has the same number of layers (6) with hidden size of 256 and 1024 for the feed forward block. The gumbel softmax layer has a temperature of 1 and $\gamma$ is set to 1.

For the IWSLT experiments, everything is the same except the hidden size of the Transformer that is set to 256. 
The fully connected layer has a hidden size of 512. 
The generator is also half the size of the transformer's hidden size.
The learning rate is adapted to the hidden size as specified in \cite{vaswani2017}. Models are trained on 1 V100 gpu with 60K steps. 


\subsection{Robustness Metrics}
\label{robustness}

To evaluate robustness, different types of noises are injected in a test set.
For instance, an unknown character is added at the beginning of the sentence, noted UNK.S or at the end, noted  UNK.E. This rare character may perturb the model and is an easily reproducible test as done in \cite{clinchant2019}. In addition, we perturb the source sentence by replacing a small proportion $p=0.1$ of words by the best candidate given by the public xlm-roberta language model. We note this test set XLM \footnote{the original words are excluded from the candidates}.
Finally, we add character level operations, noted CHAR, with some random insertions, deletions and swaps with a certain proportion
in the sentence. 

Once the perturbation is defined, 
we used the BLEU consistency noted {\bf Cy(Bleu)} \cite{niu2020}: it measures how much the translation changed compared to the initial hypothesis and can be understood as stability measure. In addition, we report $\mathbf{\Delta}(\text{chrf})$ \cite{clinchant2019}: the mean sentence level drop in chrf, it is negative and the closer to zero the best it is.
 

\begin{table*}[h]
    \centering
    \begin{tabular}{lccccccc c}
    \toprule
    \multicolumn{1}{c}{\multirow{2}{*}{\textbf{Model}}} & BLEU &  \multicolumn{2}{c}{UNK.S}  & \multicolumn{2}{c}{UNK.E}& \multicolumn{2}{c}{XLM $p=0.1$} & CHAR \\         
    
                &  \small{WMT14}  & $\Delta$(chrf) & Cy(Bleu) &  $\Delta$(chrf) & Cy(Bleu)  & $\Delta$(chrf) & Cy(Bleu)  & Cy(Bleu) \\
    \midrule     
    \small{Base 32k}          &    26.1      & -0.012  & 65.3  & -0.007   & 74.3  & -0.078  & 46.7 & 28.3 \\
    \small{Base 32k BatchX2}  &   26.7       & -0.013   & 62.4  & -0.009  & 72.8  & -0.082  & 45.9  & 27.4 \\
    \small{MAG 32k}           &  26.8        & \textbf{-0.009}  & \textbf{70.0}   &  \textbf{-0.006} & 79.7   &  \textbf{-0.071}&  \textbf{52.1} & \textbf{31.4}\\
    \small{Random 32k}       &  26.4         & -0.013  & 66.2   & -0.006   & 83.8 & -0.076   & 49.4 & 29.6 \\
    \small{MAG-NoAdv}        &  \textbf{26.9} & -0.015   & 64.8  & -0.007 & \textbf{86.8}  & -0.080  & 47.7 & 28.7 \\
    \small{MAG-NoMLM}        &  26.8    &  -0.012 & 66.7  & -0.013  & 75.9   & -0.080 &  47.6 & 28.8 \\
    \midrule 
    \small{Base 40k}          &  26.3   & -0.010    & 67.5  & -0.011   & 73.6   & -0.078   & 46.9 & 27.8  \\
    \small{Base 40k BatchX2}  &  26.4   &  -0.010 & 68.2  & -0.010   & 71.8   & -0.081  & 46.4  & 27.0 \\
    \small{MAG 40k}           &  26.7  &   \textbf{-0.005} & \textbf{71.4}  &  \textbf{-0.004}   & 79.6   &  \textbf{-0.069} & \textbf{52.3}  & \textbf{31.5} \\
    \small{Random 40k}       &  26.7   &   -0.012  &  65.4 & -0.005 & \textbf{84.6}   & -0.076  & 49.4   & 30.9\\
    \small{MAG-NoAdv}        &  26.6   &   -0.012 & 67.0  &  -0.008   & 78.2  & -0.078  & 48.6  & 29.6 \\
    \small{MAG-NoMLM}        &  \textbf{26.8}  &  -0.012 & 65.7  &-0.015  &70.9  &-0.078  & 47.9  & 28.4 \\
    
    \bottomrule
   \end{tabular}
  \caption{WMT'14 Test BLEU  (First Column) and Robustness measures on noisified test set. $\Delta$(chrf) and Consistency of BLEU ( noted Cy(Blue). UNK.S add an unknown character at the start of sentence, UNK.E at the end; XLM replace some words ; CHAR uses character level operations (cf section \ref{robustness}). $\Delta$(chrf) : the lower the better. }

   \label{tab:wmt}
\end{table*}

\subsection{IWSLT results}


On IWSLT, the base Transformer model achieved a 33.5 BLEU whereas random augmentation reached 34.0, the Doubly Adversarial 34.9 and
our MAG model 35.0 of BLEU. Additional robustness analysis and ablation studies (not included here due to lack of space) shows that:  
a) all the methods improve the robustness of the base Transformer model b) the discrimination loss brings the most benefit and the adversarial part has little effect. However, since this dataset is quite small, any regularization strategy brings improvements. 
Nevertheless, the difference between MAG and the Doubly Adversarial paper is that our model is much faster to train. The training times on a normal transformer model on IWSLT14 are around 3 hours on a single V100 GPU. 
For the doubly adversarial inputs an experiment on a V100 GPU on IWSLT14 is 12h long and for our Masked Adversarial Model, it is 5h long.

\subsection{WMT Results}

\begin{table}[h]
    \centering
    \begin{tabular}{llll}
     \toprule
    \textbf{Model}                           & \small{WMT20}    & \small{News20} & \small{News18} \\ 
                                            &  \small{Robust}   &  &  \\ \hline
    \small{Base 32k}                          &  19.4          &   22.4  &39.6 \\
    \small{Base 32k BatchX2}                 &   19.1          &  22.2  &39.8 \\
    \small{Base 40k BatchX2}                 &  19.4           &  22.8  & 39.3  \\
    \small{Random 40k}                       &  {19.0}         &   23.1 & 40.1 \\ 
    \small{MAG 32k}                          &  19.8           & 23.1  & 40.3 \\ 
    \small{MAG 40k}                          &  \textbf{19.9}  &  \textbf{23.8} & \textbf{40.4} \\
     \bottomrule
    \end{tabular}
    \caption{Out-of domain BLEUs for models trained on  WMT17}
    \label{tab:wmt_domains}
\end{table}

First of all, none of the previous papers actually compared their results to baselines with bigger batch size and a random substitution attack. Indeed, adversarial training amounts to using a bigger batch where tokens have been replaced.  Secondly, we did not manage to reproduce the results on WMT14 given in the doubly adversarial paper, even after exchanging with the author and discussing crucial bits of our code.

To better understand the results of MAG, we show the performance of MAG, where the generator is not adversarial (i.e $\gamma=0$), a model we refer to as MAG-NoAdv , and a model MAG-NoMLM where the generator is purely adversarial (no MLM loss is taken into account). Finally, we ran experiment with both a BPE size of 32K and 40K to assess the effect of subword size on performance.

Table \ref{tab:wmt} compares the results of MAG with different baselines on WMT14 test set and its robustness for the noisy test sets. First, the table shows that the model BLEUs is on par or better than the baseline.
Second, the random baseline improves robustness for most cases. $\Delta$(chrf) and Cy(Bleu)  seem to agree on the best model except for UNK.E. 
The results show that both MAG-NoAdv and MAG-NoMLM improves the results: both pseudo data augmentation and adversarial training are beneficial but, all in all, MAG is more robust according to the different metrics. Furthermore, we also tested a variant of MAG without the discrimination loss and the results were similar, showing it had no impact on this dataset on the contrary to IWSLT.

Finally, we test the domain robustness of these models on out-of domain test set in table \ref{tab:wmt_domains}: the WMT20 Robustness test set, including many noisy sentences, and two news test sets. MAG performs the best indicating better out of domain robustness. Please note that those results are lower compared to state of the art as stronger baselines use bigger training set (30M sentences compared to 4.5M sentences), ensembling, or pretraining techniques. For future work, we plan to investigate the robustness of MAG on bigger training collection, combined with inline casing, bpe-dropout and frequency sampling for masking subwords. Finally, one limitation of our model is its reliance on subwords as units of perturbation. If insertions are possible to emulate with an artificially added token, deletions are not possible yet.



	\section{Conclusion and Future work}
	Dealing with robustness is a computationally challenging task for neural machine translations  models. Recent works relied on gradient based attacks to adversarially train NMT models. To our knowledge, this work is the first  to learn a generator that captures \textit{explicitly}  a trade-off between language model augmentation and adversarial perturbations. The model is much faster to compute compared to the Doubly Adversarial method, it also improves robustness over several baselines in our experiments and BLEU score for out-of domain test sets. Finally, it  would be worth investigating MAG for others Seq2Seq tasks.

\bibliographystyle{acl_natbib}
\bibliography{report}

\end{document}